\documentclass[11pt,a4paper]{article}
\usepackage[hyperref]{acl2021}
\usepackage{url}
\usepackage{times}
\usepackage{latexsym}

\usepackage{microtype}

\makeatletter
\def\blfootnote{\xdef\@thefnmark{}\@footnotetext}
\makeatother
\usepackage{booktabs}
\usepackage{xspace}
\usepackage{tabularx}
\usepackage{wrapfig}
\usepackage{subfig}
\newcommand{\F}{F$_1$\xspace}
\newcommand{\rt}[1]{\rotatebox{90}{#1}}
\newcommand{\twocol}[1]{\multicolumn{2}{c}{#1}}
\newcommand{\onecol}[1]{\multicolumn{1}{c}{#1}}
\newcommand{\emoMEbase}{\textit{Emo-ME-Base}\xspace}
\newcommand{\emoNNbase}{\textit{Emo-NN-Base}\xspace}
\newcommand{\cpmNNbase}{\textit{Cpm-NN-Base}\xspace}
\newcommand{\cpmMEbase}{\textit{Cpm-ME-Base}\xspace}
\newcommand{\cpmMEadv}{\textit{Cpm-ME-Adv}\xspace}
\newcommand{\emocpmMEpred}{\textit{Emo-Cpm-ME-Pred}\xspace}
\newcommand{\emocpmMEgold}{\textit{Emo-Cpm-ME-Gold}\xspace}
\newcommand{\emocpmNNpred}{\textit{Emo-Cpm-NN-Pred}\xspace}
\newcommand{\emocpmNNgold}{\textit{Emo-Cpm-NN-Gold}\xspace}
\newcommand{\mtlmh}{\textit{MTL-MH}\xspace}
\newcommand{\mtlxs}{\textit{MTL-XS}\xspace}
\usepackage{scalefnt}
\usepackage{multirow}
\usepackage{graphicx}
  
\aclfinalcopy

\title{Emotion Recognition under Consideration of the\\ Emotion
  Component Process Model}

\author{Felix Casel$^*$, Amelie Heindl$^*$, and Roman Klinger \\
  Institut f\"ur Maschinelle Sprachverarbeitung, University of Stuttgart \\
  Pfaffenwaldring 5b, 70569 Stuttgart, Germany\\
  \{firstname.lastname\}@ims.uni-stuttgart.de \\
}

\date{}

\begin{document}
\maketitle
\begin{abstract}
  Emotion classification in text is typically performed with neural
  network models which learn to associate linguistic units with
  emotions. While this often leads to good predictive performance, it
  does only help to a limited degree to understand how emotions are
  communicated in various domains.  The emotion component process
  model (CPM) by \newcite{Scherer2005} is an interesting approach to
  explain emotion communication. It states that emotions are a
  coordinated process of various subcomponents, in reaction to an
  event, namely the subjective feeling, the cognitive appraisal, the
  expression, a physiological bodily reaction, and a motivational
  action tendency. We hypothesize that these components are associated
  with linguistic realizations: an emotion can be expressed by
  describing a physiological bodily reaction (``he was trembling''),
  or the expression (``she smiled''), etc. We annotate existing
  literature and Twitter emotion corpora with emotion component
  classes and find that emotions on Twitter are predominantly
  expressed by event descriptions or subjective reports of the
  feeling, while in literature, authors prefer to describe what
  characters do, and leave the interpretation to the reader. We
  further include the CPM in a multitask learning model and find that
  this supports the emotion categorization. The annotated corpora are
  available at \url{https://www.ims.uni-stuttgart.de/data/emotion}.
\end{abstract}

\blfootnote{\hspace{-0.65cm}$^*$The first two authors contributed equally to this work.}

\section{Introduction}
\label{introduction}
The task of emotion classification from written text is to map textual
units, like documents, paragraphs, or sentences, to a predefined set
of emotions.  Common class inventories rely on psychological
theories such as those proposed by \newcite{Ekman1992}
(\textit{anger}, \textit{disgust}, \textit{fear}, \textit{joy},
\textit{sadness}, \textit{surprise}) or \newcite{Plutchik2001}. Often,
emotion classification is tackled as an end-to-end learning task,
potentially informed by lexical resources (see the SemEval Shared Task
1 on Affect in Tweets for an overview of recent approaches
\cite{Mohammad2018}).

While end-to-end learning and fine-tuning of pretrained models for
classification have shown great performance improvements in contrast
to purely feature-based methods, such approaches typically neglect the
existing knowledge about emotions in psychology (which might help in
classification and to better understand how emotions are
communicated). There are only very few approaches that aim at
combining psychological theories (beyond basic emotion categories)
with emotion classification models: We are only aware of the work by
\newcite{Hofmann2020}, who incorporate the cognitive appraisal of
events, and
\newcite{Buechel2020}, who jointly learn affect (valence, arousal) and
emotion classes; next to knowledge-base-oriented modelling of events
by \newcite{Balahur2012} and \newcite{Cambria2014}.

An interesting and attractive theory for computational modelling of
emotions that has not been used in natural language processing yet is
the emotion component process model \cite[CPM]{Scherer2005}. This
model states that emotions are a coordinated process in five
subsystems, following an event that is relevant for the experiencer of
the emotion, namely a \emph{motivational action tendency}, the
\emph{motor expression} component, a \emph{neurophysiological, bodily
  symptom}, the \emph{subjective feeling}, and the \emph{cognitive
  appraisal}. The cognitive appraisal has been explored in a
fine-grained manner by \newcite{Hofmann2020}, mentioned above. The
subjective feeling component is related to the dimensions of
affect.\footnote{There exists other work that has
 been motivated by appraisal theories, but that is either rule-based
 \cite{Shaikh2009,Udochukwu2015} or does not explicitly model
 appraisal or component dimensions \cite{Balahur2012,Rashkin2018}.}

We hypothesize (and subsequently analyze) that emotions in text are
communicated in a variety of ways, and that these different stylistic
means follow the emotion component process model. The communication of
emotions can either be an explicit mention of the emotion name (``I am
angry''), focus on the motivational aspect (``He wanted to run
away.''), describe the expression (``She smiled.'', ``He shouted.'')
or a physiological bodily reaction (``she was trembling'', ``a tear
was running down his face''), the subjective feeling (``I felt so
bad.''), or, finally, describe a cognitive appraisal (``I wasn't sure
what was happening.'', ``I am not responsible.'').

With this paper, we study how emotions are communicated (following the
component model) in Tweets (based on the Twitter Emotion Corpus TEC,
by \newcite{Mohammad12}) and literature (based on the REMAN corpus by
\newcite{Kim2018}). We post-annotate a subset of 3041 instances with
the use of emotion component-based emotion communication categories,
analyze this corpus, and perform joint modelling/multi-task learning
experiments. Our research goals are (1) to understand if emotion
components are distributed similarly across emotion categories and
domains, and (2) to evaluate if informing an emotion classifier about
emotion components improves their performance (and to evaluate various
classification approaches). We find that emotion component and emotion
classification prediction interact and benefit from each other and
that emotions are communicated by means of various components in
literature and social media. The corpus is available at
\url{https://www.ims.uni-stuttgart.de/data/emotion}.

\section{Background and Related Work}
\subsection{Emotion Models}
\label{emo_models}
Emotion models can be separated into those that consider a discrete
set of categories or those that focus on underlying principles like
affect. The model of basic emotions by \newcite{Ekman1992} considers
anger, disgust, fear, joy, sadness, and surprise.  According to his
work, there are nine characteristics that a basic emotion fulfills:
These are (1) distinctive universal signals, (2) presence in other
primates, (3) distinctive physiology, (4) distinctive universals in
antecedent events, (5) coherence among emotional response, (6) quick
onset, (7) brief duration, (8) automatic appraisal, and (9) unbidden
occurrence.  His model of the six universal emotions constitutes one
of the most popular emotion sets in natural language processing.  Yet
it might be doubted if this set is sufficient.  \newcite{Plutchik2001}
proposed a model with eight main emotions, visualized on a colored
wheel. In this visualization, opposites and distance of emotion names
are supposed to correspond to their respective relation.

A complementary approach to categorizing emotions in discrete sets is
advocated by \newcite{russell1977evidence}. Their dimensional affect
model corresponds to a 3-dimensional vector space with dimensions for
pleasure-displeasure, the degree of arousal, and
dominance-submissiveness (VAD). Emotion categories correspond to
points in this vector space.
A more expressive alternative to the VAD model of affect is motivated
by the cognitive appraisal process that is part of emotions. The model
of \newcite{smith1985patterns} introduces a set of variables that they
map to the principle components of pleasantness,
responsibility/control, certainty, attention, effort, and situational
control. They show that these dimensions are more powerful to
distinguish emotion categories than VAD.

Appraisals are also part of the emotion component process model by
\newcite{Scherer2005}, which is central to this paper.  The five
components are \emph{cognitive appraisal}, \emph{neurophysiological
  bodily symptoms}, \emph{motor expressions}, \emph{motivational
  action tendencies}, and \emph{subjective feelings}. \emph{Cognitive
  appraisal} is concerned with the evaluation of an event.  The event
is assessed regarding its relevance to the individual, the
implications and consequences it might lead to, the possible ways to
cope with it and control it, and its significance according to
personal values and social norms. The component of
\emph{neurophysiological symptoms} regards automatically activated
reactions and symptoms of the body, like changes in the heartbeat or
breathing pattern.  The \emph{motor expression} component contains all
movements, facial expressions, changes concerning the speech, and
similar patterns.  Actions like attention shifts and movement with
respect to the position of the event are part of the
\emph{motivational action tendencies} component. Finally, the
component of \emph{subjective feelings} takes into account how strong,
important, and persisting the felt sensations
are. \newcite{Scherer2005} argues that it is possible to infer the
emotion a person is experiencing by analyzing the set of changes in
the five components. \newcite{Scherer2009} also points out that
computational models must not ignore emotion components.

\subsection{Emotion Analysis in Text}
The majority of modelling approaches focuses on the analysis of
fundamental emotions \citep[see][]{Alswaidan,Mohammad2018,Bostan2018} or
on the recognition of valence, arousal, and dominance
\cite{Buechel2017}. Work with a focus on other aspects of emotions
is scarce.

Noteworthy, though this has not been a computational study, is the
motivation of the ISEAR project \cite{Scherer94}, from which a textual
corpus originated, which is frequently used in NLP. It consists of
event descriptions and is therefore relevant for appraisal
theories. Further, participants in that study have not only been asked
to report on events they experienced, but they also report additional
aspects, including the existence of bodily reactions. However, their
work does not focus on the \emph{linguistic realization} of emotion
components, but on the \emph{existence} in the described event.

Similarly, \newcite{Troiano2019} asked crowdworkers to report on
events that caused an emotion. This resource has then been
postannotated with appraisal dimensions \cite{Hofmann2020}. This is
the only recent work we are aware of that models appraisal as a
component of the CPM to predict emotion categories, next to the
rule-based classification approach by \newcite{Shaikh2009}, who built
on top of the work by \newcite{Clore}. Another noteworthy related work
is SenticNet, which models event properties including people's goals,
for sentiment analysis \citep{Cambria2014}.

The only work we are aware of that studies emotion components (though
not following the CPM, and without computational modelling), is the
corpus study by \newcite{Kim2019}. They analyze if emotions in fan
fiction are communicated via facial descriptions, body posture
descriptions, the appearance, look, voice, gestures, subjective
sensations, or spatial relations of characters. This set of variables
is not the same as emotion components, however, it is related. They
find that some emotions are preferred to be described with particular
aspects by authors. Their work was motivated by the linguistic study
of \newcite{vanmeel1995}.

In contrast to their work, our study compares two different domains
(Tweets and Literature), and follows the emotion component process
model more strictly. Further, we show the use of that model for
computational emotion classification through multi-task learning.

\section{Corpus Annotation}
\subsection{Corpus Selection}
To study the relation between emotion components and emotions, we
annotate subsets from two different existing emotion corpora from two
different domains, namely literature and social media.

For literature, we use the REMAN corpus \cite{Kim2018}, which consists
of fiction written after the year 1800. It is manually annotated with
text spans related to emotions, as well as their experiencers, causes,
and targets. Emotion cue spans are annotated with the emotions of
anger, fear, trust, disgust, joy, sadness, surprise, and anticipation,
as well as `other emotion'. From the 1720 instances, we randomly
sample a subset of 1000.  Each instance comprises a sentence triple
and may contain any number of annotated spans. We map the emotions
associated to spans to the text instances as the union of all
labels, which leads to a multi-label classification
task. Instances without emotion annotations are considered
`neutral'.

For the social media domain, we choose the Twitter Emotion Corpus
(TEC) \citep{Mohammad12}.  The emotion categories are anger, disgust,
fear, joy, sadness, and surprise.  TEC consists of approximately
21,000 posts from Twitter that have a hashtag at the end which states
one of the six mentioned emotions. According to the authors, the
validity of hashtags as classification labels is commensurable to the
inter-annotator agreements of human annotators. We randomly sample
2041 instances with the emotion hashtags as labels for the creation of
our corpus.  Each instance equals one post and has exactly one emotion
label.

\begin{table*}[t]
  \centering\small
  \begin{tabularx}{\linewidth}{lp{0.33\textwidth}X}
    \toprule
    Component&  Explanation of Example & Example \\
    \cmidrule(r){1-1}\cmidrule(lr){2-2}\cmidrule(l){3-3}
    Cognitive appraisal
             &
               evaluation of the pleasantness of an event.
                           &
                             Thinks that @melbahughes had a great 50th birthday party
    \\
    Neurophysiol.\ symptoms
             &
               change in someone's heartbeat.
                           &
                             Loves when a song makes your heart race [...]
    \\
    Motiv.\ Action tendencies
             &
               urge to attack a person or object.
                           &
                             sometimes when i think bout you i want to beat the shit out of your face so everyone can see how ugly you are inside and out
    \\
    Motor expressions
             &
               facial expression.
                           &
                             @TheBodyShopUK when I walk in the room and my 9month old nephew recognises me and his face lights up with the biggest smile thats 100\%
    \\
    Subjective feelings
             &
               internal feeling state.
                           &
                             Feelin a bit sad tonight
    \\
    \bottomrule
  \end{tabularx}
  \caption{Excerpt of the final annotation guidelines including examples from TEC.\\[-1.6mm]}
  \label{tab:guidelines}
\end{table*}

\begin{table}[t]
  \centering\small
  \begin{tabular}{lcc}
    \toprule
    Component  & round 1 &   round 2  \\
    \cmidrule(r){1-1}\cmidrule(lr){2-2}\cmidrule(l){3-3}
    Cognitive appraisal & 0.288 & 0.777 \\
    Neurophysiological symptoms & 0.459 & -- \\
    Motiv.\ Action tendencies & 0.444 & 0.732\\
    Motor expressions & 0.643 & 0.617 \\
    Subjective feelings & 0.733 & 0.793\\
    \bottomrule
  \end{tabular}	
  \caption{Inter-annotator agreement after the different annotation
    rounds during the guideline creation process measured with Cohen's
    $\kappa$. In the second round, no annotator detected the neurophysiological
    component in the sample instances.}
  \label{tab:annotatorAgreement}
\end{table}

\subsection{Annotation Procedure and Inter-Annotator Agreement}
We annotate the emotion component dimensions independently: The
existence of a CPM label means that this component is mentioned
somewhere in the text, independent of its function to communicate one
of the emotions. This is a simplification due to the fact that it
turned out to be difficult to infer from the limited context of an
instance if an emotion category and an
emotion component mention are actually in relation. Further, this
procedure also ensures that there is no information leak introduced in
the annotation process (e.g., that components are only annotated if
they indeed inform the emotion, and that a model could learn from its
sheer presence).

We refined the annotation guidelines in an iterative process with two
annotators.  Annotator 1 is a 23 year-old female undergraduate
computer science student, Annotator 2 is a 28 year-old male graduate
student of computational linguistics. We first defined a list of
guidelines for each emotion component, then let each annotator label
40 randomly sampled instances (20 each in two iterations) out of each
corpus and measured the inter-annotator agreement. Based on instances
with disagreement, we refined the guidelines. The achieved
inter-annotator agreement scores are displayed in
Table~\ref{tab:annotatorAgreement}. We observe that particularly the
concepts of cognitive appraisal and motivational action tendencies
have been clarified.  During this process, for example, the discussion
of the instance \textit{``He did so, and to his surprise, found that
  all the bank stock had been sold, and transferred''} lead to the
addition of a rule stating that the explicit mention of a feeling has
to be annotated with subjective feeling.  A rule for the annotation of
tiredness as neurophysiological symptoms was created due to the
instance \textit{``Here he remained the whole night, feeling very
  tired and sorrowful.''}.  Concerning the annotation of verbal
communication as motor expression, we decided to only annotate
instances with verbal communications that address an emotional
reaction or instances with interjections as for example `oh' or `wow'.
With this clarification, the instance \textit{```Jolly rum thing about
  that boat,' said the spokesman of the party, as the boys continued
  their walk. `I expect it got adrift somehow,' said another. `I don't
  know,' said the first.''} should not be annotated, whereas
\textit{```Sounds delightful.' `Oh, it was actually pretty cool.'{}''}
should (this aspect has particularly appeared in the second annotator
training round, which lead to a slight decrease in agreement).  We
make the annotation guidelines available together with our
corpus. Table~\ref{tab:guidelines} shows a short excerpt.

After the refinement process concluded, Annotator 1
annotated the subsample of TEC and Annotator 2 annotated the subsample
of REMAN.

\subsection{Corpus Statistics}
We show corpus statistics in Table~\ref{tab:stats} to develop an
understanding how emotions are communicated in the two domains.  For both
corpora, we observe that cognitive appraisal is most frequent.  In
TEC, the second most dominant component is subjective feeling, while
in REMAN it is the motor expression.  The amount of subjective feeling
descriptions is substantially lower for literature than for social
media -- which is in line with the show-don't-tell paradigm which is
obviously not followed in social media as it is in literature.

Components are not distributed equally across emotions.  Particularly
noteworthy is the co-occurrence of disgust with neurophysiological
symptoms in social media, but not in literature where this component
dominates the emotion of fear.  We also observe a particularly high
co-occurrence of the subjective feeling component with fear for social
media, which is not the case for literature.  In literature, the
motivational action tendency component co-occurs with anger (and
anticipation) more frequently than with all other emotions.  This is
not the case for the social media domain.  On the REMAN corpus,
components occur least frequently when there is no emotion across all
components.  For both corpora, neurophysiological symptoms make up the
smallest share of components, even more so in the case of TEC than
REMAN.

In a comparison of social media and literature, we
observe that emotions are distributed more uniformly in literature.
The relative number of co-occurrences of CPM components with emotions
varies more for REMAN than for the TEC corpus.

\begin{table*}
  \newcommand{\sep}{\cmidrule(r){2-2}\cmidrule(lr){3-4}\cmidrule(lr){5-6}\cmidrule(lr){7-8}\cmidrule(lr){9-10}\cmidrule(lr){11-12}\cmidrule(r){13-13}}
  \newcommand{\sepp}{\cmidrule(lr){1-4}\cmidrule(lr){5-6}\cmidrule(lr){7-8}\cmidrule(lr){9-10}\cmidrule(lr){11-12}\cmidrule(r){13-13}}
  \centering\small
  \setlength{\tabcolsep}{8pt}
  \renewcommand{\arraystretch}{1.05}
  \begin{tabular}{ll rr rr rr rr rr r}
    \toprule
    & Emotion&
\twocol{Cognitive} & \twocol{Phys.} & \twocol{Motiv.\ Action} & \twocol{Motor Exp.} & \twocol{Subject.} & \onecol{Total} \\
    \sep
    \multirow{7}{*}{\rt{TEC}}
    &Anger    & 127 & (75\%) & 8  & (5\%)  & 30  & (18\%) & 20 & (12\%) & 49  & (29\%) & 169 \\
    &Disgust  & 65  & (83\%) & 11 & (14\%) & 6   & (8\%)  & 17 & (22\%) & 19  & (24\%) & 78 \\
    &Joy      & 606 & (71\%) & 59 & (7\%)  & 176 & (21\%) & 95 & (11\%) & 233 & (27\%) & 848 \\
    &Sadness  & 323 & (87\%) & 13 & (3\%)  & 58  & (16\%) & 53 & (14\%) & 142 & (38\%) & 373 \\
    &Fear     & 196 & (74\%) & 9  & (3\%)  & 37  & (14\%) & 27 & (10\%) & 130 & (49\%) & 266 \\
    &Surprise & 219 & (71\%) & 2  & (1\%)  & 55  & (18\%) & 55 & (18\%) & 83  & (27\%) & 307 \\
    \sep
    &Total.   & 1536 & (75\%) & 102 &(5\%)& 362 &(18\%)& 267 &(13\%)& 656 &(32\%)&\\
    \sepp
    \multirow{11}{*}{\rt{REMAN}}
    &Anger    & 66  & (67\%) & 7  & (7\%)  & 40  & (41\%) & 61  & (62\%) & 25  & (26\%) & 98 \\ 
    &Anticip. & 69  & (59\%) & 6  & (5\%)  & 50  & (43\%) & 63  & (54\%) & 19  & (16\%) & 117 \\
    &Disgust  & 81  & (86\%) & 5  & (5\%)  & 21  & (22\%) & 33  & (35\%) & 16  & (17\%) & 94 \\ 
    &Fear     & 96  & (67\%) & 33 & (23\%) & 35  & (24\%) & 70  & (49\%) & 34  & (24\%) & 143 \\ 
    &Joy      & 121 & (57\%) & 11 & (5\%)  & 28  & (13\%) & 117 & (55\%) & 66  & (31\%) &  213 \\ 
    &Neutral  & 39  & (34\%) & 0  & (0\%)  & 13  & (11\%) & 22  & (19\%) & 3   & (3\%) & 116  \\ 
    &Other    & 64  & (57\%) & 11 & (10\%) & 21  & (19\%) & 53  & (47\%) & 21  & (19\%) & 113  \\ 
    &Sadness  & 94  & (69\%) & 19 & (14\%) & 22  & (16\%) & 66  & (49\%) & 42  & (31\%) & 136  \\ 
    &Surprise & 103 & (74\%) & 11 & (8\%)  & 21  & (15\%) & 83  & (60\%) & 22  & (16\%) & 139 \\ 
    &Trust    & 94  & (82\%) & 2  & (2\%)  & 17  & (15\%) & 34  & (30\%) & 27  & (23\%) & 115  \\ 
    \sep
    &Total& 610 &(61\%)& 76 &(8\%)&	 190 &(19\%)&	 440 &(44\%)&	 174 &(17\%)& \\
    \bottomrule
  \end{tabular}%
  \caption{Total/relative counts of CPM components and
    emotions in our reannoted TEC and REMAN subsamples. Note that the
    CPM categorization is a multi-label task, with 1000 instances in
    REMAN and 2041 instances reannotated in TEC.}
  \label{tab:stats}
\end{table*}

\section{Methods}
We will now turn to the computational modelling of emotion components
and evaluate their usefulness for emotion classification. We evaluate
a set of different feature-based and deep-learning based
classification approaches to join the tasks of emotion classification
and component classification.

\subsection{Emotion Classifier}
As baseline emotion classification models which are not particularly
informed about components, we use two models: \textbf{\emoMEbase} is a
maximum entropy (ME) classifier with TF-IDF-weighted bag-of-words unigram
and bigram features. As preprocessing, we convert all words to
lowercase, and stem them with the PorterStemmer.  On
TEC, with its single-label annotation, \emoMEbase consists of one
model, while on REMAN with multi-label annotation, we use 10
binary classifiers.

\begin{figure}[t]
  \centering
  \subfloat[][\textit{Emo/CPM-NN-Base}]{\includegraphics[scale=0.5,page=1]{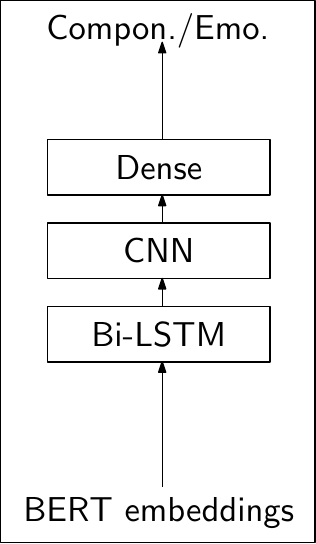}\label{fig:NN-base}}
  \hfill
  \subfloat[][\hbox{\emocpmNNpred}]{\includegraphics[scale=0.5,page=2]{nnmodels}\label{fig:NN-pipe}}
  \hfill
  \subfloat[][\mtlxs]{\includegraphics[scale=0.5,page=3]{nnmodels}\label{fig:NN-XS}}
  \caption{Neural Model Architectures (subset)}
  \label{fig:nnarch}
\end{figure}

Our neural baseline \textbf{\emoNNbase} uses pre-trained BERT sentence
embeddings%
\footnote{https://tfhub.dev/google/experts/bert/wiki\_books/sst2/1}
\citep{devlin2018bert} as input features. Inspired by
\newcite{Chen2018,Sosa2017}, the network architecture consists of a
bidirectional LSTM layer \citep{hochreiter1997long}, followed by a
convolutional layer with kernel sizes 2, 3, 5, 7, 13, and 25.  The
outputs of the convolutional layer are max-pooled over the dimension
of the input sequence, inspired by \newcite{collobert2011natural}.
Stacked on top of the pooling layer is a fully connected layer.  Its
outputs are finally fed into an output layer with a sigmoid activation
function (see Figure~\ref{fig:NN-base}).\footnote{We selected this
  architecture based on preliminary experiments on the validation
  data. We evaluated it against LSTM-Dense Layer and CNN-LSTM
  architectures.}

We use dropout regularization after each layer.  The network uses a
weighted cross-entropy loss function, whereby the loss of false
negatives is multiplied by 4 to increase recall.  The model is trained
using an Adam optimizer \citep{kingma2017adam}.  All network
parameters of this model and subsequent neural models are determined
using a subset of the training data as development set for the REMAN
corpus and using 10-fold cross-validation for the TEC corpus. Details
of the resulting hyperparameters are listed in the Appendix.

\subsection{Component Classifier}
The emotion component classifiers predict which of the five CPM
components occur in a text instance.
Our \textbf{\cpmMEbase} baseline models (one for each component) only use
bag-of-words features in the same configuration as
\emoMEbase.

In the model \textbf{\cpmMEadv}, we add task-specific features, namely
features derived from manually crafted small dictionaries with words
associated with the different components.  Those dictionaries were
developed without considering the corpora and with inspiration from
\newcite{Scherer2005} and contain on average 26 items.  Further, we
add part-of-speech tags (calculated with
spaCy\footnote{https://spacy.io/usage/linguistic-features\#pos-tagging},
\newcite{spacy}) and glove-twitter-100
embeddings\footnote{https://nlp.stanford.edu/projects/glove/}
\citep{pennington-etal-2014-glove}.  Additionally, only for the
cognitive appraisal component, we run the appraisal classifier
developed by \newcite{Hofmann2020} and use the predictions as
features.\kern-3pt\footnote{http://www.ims.uni-stuttgart.de/data/appraisalemotion}
For each component individually, the best-performing combination of
these features is chosen.

The \textbf{\cpmNNbase} is configured analogously to \emoNNbase.
The primary reason for using an equivalent setup is to
facilitate a multi-head architecture as joint model for both tasks in
the next step.

\subsection{Joint Modelling and Multi-Task Learning of Emotions and Components}
\label{sec:joint_models}
To analyze if emotion classification benefits from the component
prediction (and partially also vice versa), we set up several model
configurations.

In \textbf{\emocpmMEpred}, we predict the emotion with \cpmMEadv and
use these predictions as features. Other than that, \emocpmMEpred
corresponds to \emoMEbase. In \textbf{\emocpmMEgold}, we replace the
predictions by gold component annotations to analyze error
propagation.

\textbf{\emocpmNNpred} and \textbf{\emocpmNNgold} are configured
analogously and follow the same architecture as \emoNNbase with the
following differences: A binary vector with the CPM annotations is
introduced as additional input feature, feeding into a fully connected
layer.  Its outputs are concatenated with the outputs of the
penultimate layer and passed to another fully connected layer,
followed by the output layer.

\emocpmNNpred uses \cpmNNbase to obtain component predictions, but the
weights of \cpmNNbase are frozen.  The basic network architecture
resembles that of the \emocpmNNgold model, replacing the
additional CPM input vector with the \cpmNNbase model (see Figure
\ref{fig:NN-pipe}).  Its outputs are, again, fed into a fully connected
layer which is connected to the output layer.

Next to the models that make use of the output of the CPM classifiers
for prediction, we use two multi-task learning models which predict
emotions and components based on shared latent variables. For a
multi-head variant (\textbf{\mtlmh}), the basic architectures of the
individual models for both tasks remain the same.  Outputs of the CNN
layer are fed to two separate, task-specific, fully connected layers.
This model has two output layers, one for emotion classification and
one for CPM component classification.  Both tasks use the weighted
cross entropy loss function to increase recall.

\begin{table*}[t]
  \centering\small
  \renewcommand{\arraystretch}{1.15}
	\begin{tabular}{ll ccc ccc ccc ccc ccc}
	\toprule
	&
	& \multicolumn{3}{c}{\cpmMEbase}
	& \multicolumn{3}{c}{\cpmMEadv}
	& \multicolumn{3}{c}{\cpmNNbase}
	& \multicolumn{3}{c}{\mtlxs}
	& \multicolumn{3}{c}{\mtlmh}
	\\
	\cmidrule(r){2-2}\cmidrule(lr){3-5}\cmidrule(rl){6-8}\cmidrule(rl){9-11}\cmidrule(rl){12-14}\cmidrule(rl){15-17}
	& Component  & P & R& \F & P & R & \F & P & R & \F & P & R & \F & P & R & \F \\
	\cmidrule(r){2-2}\cmidrule(lr){3-5}\cmidrule(rl){6-8}\cmidrule(rl){9-11}\cmidrule(rl){12-14}\cmidrule(rl){15-17}
	\multirow{6}{*}{\rt{REMAN}}
	&Cognitive appraisal    & 60 & 98 & \textbf{75} & 60 & 98 & \textbf{75} & 60 & 98 & \textbf{75} & 60 & 98 & \textbf{75}  & 59 & 96 & 73 \\
	&Neurophysiological symp.   & 50 & 20 & 29 & 50 & 40 & \textbf{44} & 20 & 20 & 20 & 25 & 20 & 22 & 0 & 0 & 0\\
	&Motiv.\ action tendencies  &  36 & 47 & 41 & 46 & 68 &\textbf{55} & 42 & 26 & 32 & 29 & 42 & 34 & 25 & 68 & 36\\
	&Motor expressions     &  67 & 56 & 61 & 76 & 65 & \textbf{70} & 92 & 53 & 68 & 76 & 60 & 68   & 81 & 60 & 69 \\
	&Subjective feelings    &  38 & 32 & 34 & 45 & 53 & 49 & 58 & 37 & 45 & 48 & 53 & \textbf{50} & 35 & 32 & 33  \\
	\cmidrule(r){2-2}\cmidrule(lr){3-5}\cmidrule(rl){6-8}\cmidrule(rl){9-11}\cmidrule(rl){12-14}\cmidrule(rl){15-17}
	&Macro avg.&  50 &  51 & 48 & 56 & 65 & \textbf{59} & 54 & 47 & 48 & 48 & 55 & 50 & 40 & 51 & 42\\
	& Micro avg. & & &  61 & & & \textbf{67} &&& 63 &&& 63 &&& 57\\
	\midrule
	\multirow{6}{*}{\rt{TEC}}
	&Cognitive appraisal    &   72 & 99 & 84 & 76 & 98 & \textbf{86} & 76 & 88 & 81  & 77 & 90 & 83  & 75 & 91 & 82 \\
	&Neurophysiological sympt.   & 17 & 17 & 17 & 15 & 33 & 21 & 25 & 17 & 20   & 17 & 17 & 17 & 100 & 17 & \textbf{29}  \\
	&Motiv.\ action tendencies  &  42 & 57 & 48 & 50 & 74 & \textbf{60} & 46 & 51 & 49 & 48 & 57 & 52  & 45 & 54 & 49 \\
	&Motor expressions     & 47 & 52 & 49 & 41 & 61 & 49 & 55 & 58 & \textbf{56} & 50 & 48 & 49 & 62 & 32 & 43 \\
	&Subjective feelings    & 63 & 70 & 66 & 63 & 70 & 66 &74 & 81 & \textbf{77} & 61 & 81 & 69 & 57 & 80 & 67 \\
	\cmidrule(r){2-2}\cmidrule(lr){3-5}\cmidrule(rl){6-8}\cmidrule(rl){9-11}\cmidrule(rl){12-14}\cmidrule(rl){15-17}
	&Macro avg. &  48 & 59  &53 & 49 & 67 & 56 & 55 & 59 & \textbf{57} & 51 & 59 & 54  & 68 & 55 & 54 \\
	& Micro avg. &&& 70&&& 71 &&& \textbf{73} &&& 71 &&& 70 \\
	\bottomrule
\end{tabular}
  \caption{Performance of the emotion component detection
    models (multiplied by 100).}
  \label{tab:results_components}
\end{table*}

Based on the model proposed by \newcite{misra2016cross}, we use
cross-stitch units in our model \textbf{\mtlxs}. This model employs
two separate parallel instances of the \cpmNNbase architecture
introduced above, one for the CPM classification task and one for
emotion classification. 
The model additionally employs one
cross-stitch unit after the respective CNN layers.  This sharing unit
learns a linear combination of the pooled task-specific CNN activation
maps which is then passed to the task-specific fully connected layers.
The cross-stitch unit learns during training which information to
share across tasks (see Figure \ref{fig:NN-XS}).

\section{Results}
For our experiments, we use our reannotated subsample of TEC and REMAN
(not all instances available in TEC and REMAN). We split the corpora
into 90\% for training and 10\% to test.
\subsection{Component Prediction}
We start the discussion of the results with the component
classification, a classification task that has not been addressed
before and for which our data set is the first that becomes available
to the research community. Table~\ref{tab:results_components} shows
the results.

The model performances are acceptable. Macro-average \F scores on
REMAN range from .42 of \mtlmh to .59 for \cpmMEadv, and from .53
(\cpmMEbase) to .57 (\cpmNNbase) on TEC. There are, however, differences
for the components: On TEC, there are difficulties in predicting
neurophysiological symptoms. The addition of task-specific features in
\cpmMEadv shows a clear improvement across all components.

The neural baseline \cpmNNbase outperforms \cpmMEadv on TEC, and does
so without feature engineering. On REMAN, the feature-based model is
superior which might be due to the engineered features being more
commonly represented in the literature domain than in social media.
This is partially leveraged in the \mtlxs model on REMAN.

The components are not equally difficult to predict; the relations
between the components are comparable across models. The lowest
performance scores are observed for neurophysiological symptoms. This
holds across models and corpora.  For the neurophysiological component
on the literature domain, however, the engineered features in
\cpmMEadv show substantial improvement, yielding an \F score of 0.44.
Cognitive appraisal shows best prediction performances, with \F
between .73 and .86.  For TEC, we observe a correlation between
performance and class size for all components.

For REMAN, \cpmMEadv is the best-performing model. \cpmMEadv's macro
average \F of 0.59 is 9pp higher than the second best
\F-score.
For TEC, the best results are achieved
by \cpmNNbase with a macro \F of 0.57.

\begin{table*}[t]
  \centering\small
  \setlength{\tabcolsep}{8pt}
	\begin{tabular}{l|lrrrrrrrrrr|rr}
	\toprule
	\multicolumn{1}{c}{}&Model & \rt{Anger} & \rt{Anticip} & \rt{Disgust} & \rt{Fear} & \rt{Joy} & \rt{Neutral} & \rt{Other} & \rt{Sadness} & \rt{Surpr.} & \rt{Trust} & \rt{Macavg.} & \rt{Micavg. }\\
	\cmidrule(r){2-2}\cmidrule(l){3-14}
	\multirow{8}{*}{\rt{REMAN}}
	&\emoMEbase    & 0  & 0 & 0   & 0  & 0  & 0  & 0  & 0  & 0  & 0  & 0  & 0 \\
	&\emocpmMEgold & 18 & 0 & 0   & 25 & 16 & 62 & 0  & 0  & 0  & 0  & 12 & 14\\
	&\emocpmMEpred & 0  & 0 & 0   & 12 & 15 & 0  & 0  & 0  & 0  & 14 & 4  & 6\\
	&\emoNNbase    & 36 & 18 & 29 & 41 & 59 & 46 & 14 & 36 & \textbf{71} & 50 & 40 & 43\\
	&\emocpmNNgold & 56 & 22 & 28 & 37 & 68 & 71 & 15 & 39 & 50 & 60 & 45 & 45\\
	&\emocpmNNpred & 32 & 0  & \textbf{33} & 34 & \textbf{71} & 40 & 17 & \textbf{52} & 58 & 42 & 38 & 43\\
	&\mtlmh        & 35 & 16 & 24 & 39 & 62 & 49 & 22 & 48 & 67 & \textbf{56} & 42 & 42\\
	&\mtlxs        & \textbf{38} & \textbf{24} & 26 & \textbf{47} & 64 & \textbf{54} & \textbf{37} & 48 & 64 & 55 & \textbf{46} & \textbf{47}\\
	\cmidrule(r){2-2}\cmidrule(l){3-14}
	\multirow{9}{*}{\rt{TEC}}
	&\emoMEbase    & 11 &    & 0  & 53 & 64 &  &  & 43 & 38 &  & 35 & 54\\
	&\emocpmMEgold & 11  &    & 0  & 59 & 66 &  &  & 40 & 43 &  & 36 & 55\\
	&\emocpmMEpred & 11  &    & 0  & 59 & 67 &  &  & 43 & 43 &  & 37 & 55\\
	&\emoNNbase    & \textbf{41} &    & 44 & 56 & 69 &  &  & 51 & 39 &  & 50 & 57\\
	&\emocpmNNgold & 52 &    & 33 & 67 & 72 &  &  & 60 & 47 &  & 55 & 62\\
	&\emocpmNNpred & 32 &    & 0  & 59 & 70 &  &  & 53 & 44 &  & 43 & 56\\
	&\mtlmh        & 17 &    & \textbf{57} & 53 & \textbf{76} &  &  & 53 & \textbf{45} &  & 50 & 58\\
	&\mtlxs        & 34 &    & 50 & \textbf{60} & 73 &  &  & \textbf{57} & 44 &  & \textbf{53} & \textbf{61}\\
	\bottomrule
\end{tabular}
  \caption{\F (/100) results across models and emotion
    categories.
    (empty cells denote that this category
    is not available in the respective corpus. The best scores (except
    the gold setting) are printed bold face.}
  \label{tab:emoresults}
\end{table*}

\subsection{Emotion Classification}
In this section, we discuss the performance of our emotion
classification models across different configurations. One question is
how providing component information to them helps
most. Table~\ref{tab:emoresults} shows the results for all experiments.

The comparison of \emoMEbase and \emoNNbase reveals that a pure
word-based model is not able to categorize emotions in REMAN, due to
the imbalancedness in this multilabel classification setup. This
observation is in line with previous results \citep{Kim2018}. The use
of BERT's contextualized sentence embeddings leads to a strong
improvement of 43pp (against a 0 \F for \emoMEbase). The performance
of the ME models is comparably limited also on TEC, though this is
less obvious on the micro-averaged \F due to the imbalancedness of the
resource (.35 macro, .54 micro \F).

Our main research question is if emotion components help emotion
classification. In our first attempt to include this information as
features, we see some improvement. On REMAN, \emocpmMEpred ``boosts''
from 0 to 6 \F, on TEC we observe an improvement by 1pp, to .55 \F. The
inclusion of predicted component information as features in the neural
network model shows no improvement on REMAN or on TEC.

To answer the question if this limited improvement is only due to a
limited performance of the component classification model, we compare
these results to a setting, in which the predicted values are replaced
by gold labels from the annotation. This setup does show an
improvement with \emocpmMEgold to .14 \F on REMAN, which is obviously
still very low; and no improvement on TEC. However, with our neural
model \emocpmNNgold, we see the potential of gold information
increasing the score for emotion classification to .45 \F on REMAN and
.62 \F on TEC.

This is an unrealistic setting -- the classifier does not have access
to annotated labels in real world applications. However, in the
(realistic) cross-stitch multi-task learning setting of \mtlxs, we
observe further improvements: On REMAN, we achieve .47 \F (which is
even slightly higher than with gold component labels), which
constitutes an achieved improvement by 4pp to the emotion classifier
which is not informed about components. On TEC, we achieve .61 \F,
which is close to the model that has access to gold components
(.62). This is an improvement of 4pp as well in comparison to the
model that has no access to components but follows the same
architecture.

Particularly, we observe that models with component information
perform better across all emotions, with the exception of surprise on
the REMAN corpus and anger on the TEC corpus.  We can therefore
conclude that emotion component information does contribute to emotion
classification; the best-performing combination is via a cross-stitch
model.

A detailed discussion based on example predictions of the various
models is available in the Appendix.

\section{Conclusion and Future Work}
We presented the first data sets (based on existing emotion corpora)
with emotion component annotation. While \newcite{Hofmann2020} has
proposed to use the cognitive appraisal for emotion classification,
they did not succeed to present models that actually benefit in
emotion classification performance. That might be due to the fact that
cognitive appraisal classification itself is challenging, and that
they did not compare multiple multi-task learning
approaches.

With this paper we moved to another psychological theory, namely the
emotion component process model, and make the first annotations
available that closely follow this theory. Based on this resource, we
have shown that, even with a comparably limited data set size, emotion
components contribute to emotion classification. We expect that with a
larger corpus the improvement would be more substantial than it is
already now. A manual introspection of the data instances also shows
that the components indeed help. Further, we have seen that emotions
are communicated quite differently in the two domains, which is an
explanation why emotion classification systems (up-to-today) need to
be developed particularly for domains of interest.  We propose that
future work analyzes further which information is relevant and should
be shared across these tasks in multi-task learning models.

Further, we propose that larger corpora should be created across
more domains, and also that multi-task learning is not only
performed individually, but also across corpora. Presumably, the
component information in different domains is not the same, but
might be helpful across them.

\section*{Acknowledgments}
This work was supported by Deutsche Forschungsgemeinschaft (project
CEAT, KL 2869/1-2).

\section*{Ethical Considerations}
We did not collect a new data set from individuals, but did reannotate
existing and publicly available resources. Therefore, this paper does
not pose ethical questions regarding data collection.

However, emotion analysis has the principled potential to be misused,
and researchers need to be aware that their findings (though they are
not in themselves harmful) might lead to software that can do harm. We
assume that sentiment and emotion analysis are sufficiently well-known
that users of social media might be aware that their data could be
automatically analyzed. However, we propose that no automatic system
ever does report back analyses of individuals and instead does
aggregate data of anonymized posts. We do not assume that analyzing
literature data poses any risk.

One aspect of our work we would like to point out is that, in contrast
to other and previous emotion analysis research, we focus and enable
particularly the analysis of implicit (and perhaps even unconcious)
communication of emotions. That might further mean that authors of
posts in social media are not aware that their emotional state could
be computationally analyzed, potentially, they are not even fully
aware of their own affective state. We would like to point out that
automatically analyzing social media data without the explicit consent
of the users is unethical at least when the user can be identified or
identify themselves, particularly if they might not be aware of the
details of an analysis system.

\bibliography{lit}
\bibliographystyle{acl_natbib}

\clearpage

\onecolumn

\appendix

\section{Ablation Study for Feature Based Maximum Entropy
  Classification Model of Emotion Components}

Table~\ref{tab:cpm_me_adv_features} shows the performance scores if
just one additional feature is enabled (while bag-of-words always
remains available). It can be seen, that the most
advantageous feature are word embeddings.  On REMAN, \cpmMEadv
achieves a macro F1-score of 0.59 and a micro F$_1$-score of 0.67.  On
TEC, we have respective values of 0.56 and 0.71, with the high micro
score resulting from cognitive appraisal being the best performing
class while also being more than twice as frequent as any other
component.

\mbox{}\\[5mm]
\begingroup
\centering\small
\setlength{\tabcolsep}{5pt}
\renewcommand{\arraystretch}{0.9}
  \begin{tabular}{ll ccc ccc ccc ccc ccc}
    \toprule
    &
   	& \multicolumn{3}{c}{\textit{Emo-ME-Base}}
    & \multicolumn{3}{c}{Dictionaries}
    & \multicolumn{3}{c}{POS-tags}
    & \multicolumn{3}{c}{Embeddings}
    & \multicolumn{3}{c}{Appraisal prediction}
    \\
    \cmidrule(r){2-2}\cmidrule(lr){3-5}\cmidrule(rl){6-8}\cmidrule(rl){9-11}\cmidrule(rl){12-14} \cmidrule(rl){15-17}
    & Component  & P & R& \F & P & R & \F & P & R & \F & P & R & \F& P & R & \F \\
    \cmidrule(r){2-2}\cmidrule(lr){3-5}\cmidrule(rl){6-8}\cmidrule(rl){9-11}\cmidrule(rl){12-14}\cmidrule(rl){15-17}
    \multirow{6}{*}{\rotatebox{90}{REMAN}}
    &Cognitive appraisal& 60 & 98 & 75 & 60 &98 & {75} & 57 & 73 & 64& 60 & 88 & {72}& 60 & 98 &75   \\
    &Neurophysiological symptoms & 50 & 20 & 29 & 25 & 20 & 22 & 00 & 00 & 00& 40 & 40 &40& 50 & 20 & 29 \\
    &Action tendencies &  36 & 47 & 41& 38 & 42 & 40& 28 & 47 &35& 45 & 68 & {54} & 36 & 47 & 41 \\
    &Motor expressions&  67 & 56 & 61 & 68 & 58 &63&  61 & 63 & 62& 76 &65 &70& 67 & 56 & 61  \\
    &Subjective feelings&  38 & 32 & 34& 44 & 37 & 40 & 32 & 37& 34 & 45 & 53 & {49}&38 & 32 & 34\\
    \cmidrule(r){2-2}\cmidrule(lr){3-5}\cmidrule(rl){6-8}\cmidrule(rl){9-11}\cmidrule(rl){12-14}\cmidrule(rl){15-17}
    &Macro avg.&  50 &  51 & 48& 47 & 51 & 48 & 36 & 44 & 39& 53 &63 & 57&50 &  51 & 48\\
    & Micro avg. & & &  61&&& 62 &&& 52&&& 65&&&61\\
    \midrule
    \multirow{6}{*}{\rotatebox{90}{TEC}}
    &Cognitive appraisal&   72 & 99 & 84 & 72 & 99 &83&74 & 98 & 84 & 76 &97 &85& 72 & 99 &84 \\
    &Neurophysiological symptoms& 17 & 17 & 17& 11 & 17 & 13 & 00 & 00 & 00 & 12 & 33 & 17&17 & 17 & 17 \\
    &Action tendencies &  42 & 57 & 48& 40 & 51 & 45 & 42 & 63 & 50& 45 & 66 & 53&42 & 57 & 48\\
    &Motor expressions& 47 & 52 & 49& 43 & 48 & 45 & 34 & 45 & 39& 40 & 61 & 48& 47 & 52 & 49\\
    &Subjective feelings& 63 & 70 & 66& 62 & 68 & 65 & 62 & 65 & 64& 58 & 65 & 61 & 63 & 70 & 66\\
    \cmidrule(r){2-2}\cmidrule(lr){3-5}\cmidrule(rl){6-8}\cmidrule(rl){9-11}\cmidrule(rl){12-14}\cmidrule(rl){15-17}
    &Macro avg.&  48 & 59  &53& 46 & 57 & 50 & 42 & 54 & 47& 46 & 64& {53}&  48 & 59  &53\\
    & Micro avg.&&& 70&&& 69 &&& 68&&& 69&&& 70\\
    \bottomrule
  \end{tabular}
  \captionof{table}{Overview over the single feature's impact in classification
    with \cpmMEadv. Each column displays the classification results if
    only this column's feature is additionally to bag-of-words features, enabled. In the last column, the additional feature is only used for the prediction of cognitive appraisal, due to the classification assumption that the components can appear individually of each other in text.
  \label{tab:cpm_me_adv_features}}
\endgroup

\section{Detailed Emotion Results for Emotion Classification}
The results table in the main paper did, for space reasons, only show
\F scores.
Table~\ref{tab:results_neural} present the complete results for the
neural network, including precision and recall values.

\vspace{10mm}

\begingroup
\centering\small
\renewcommand{\arraystretch}{0.9}
  \begin{tabular}{ll ccc ccc ccc ccc ccc}
    \toprule
    &
    & \multicolumn{3}{c}{\emoNNbase}
    & \multicolumn{3}{c}{\emocpmNNgold}
    & \multicolumn{3}{c}{\emocpmNNpred}
    & \multicolumn{3}{c}{\mtlmh}
    & \multicolumn{3}{c}{\mtlxs}
    \\
    \cmidrule(r){2-2}\cmidrule(lr){3-5}\cmidrule(rl){6-8}\cmidrule(rl){9-11}\cmidrule(rl){12-14}\cmidrule(rl){15-17}
    & Emotion  & P & R& \F & P & R & \F & P & R & \F & P & R & \F & P & R & \F \\
    \cmidrule(r){2-2}\cmidrule(lr){3-5}\cmidrule(rl){6-8}\cmidrule(rl){9-11}\cmidrule(rl){12-14}\cmidrule(rl){15-17}
    \multirow{10}{*}{\rotatebox{90}{REMAN}}
    &Anger         & 28 & 50 & 36  & 47 & 70 & \textbf{56} & 33 & 30 & 32  & 31 & 40 & 35 & 31 & 50 & 38\\
    &Anticipation  & 18 & 18 & 18  & 19 & 27 & 22 & 0 & 0 & 0  & 12 & 27 & 16   & 17 & 36 & \textbf{24}\\
    &Disgust       & 20 & 56 & 29 & 20 & 44 & 28 & 24 & 56 & \textbf{33}  & 16 & 56 & 24  & 18 & 44 & 26\\
    &Fear          & 35 & 50 & 41  & 25 & 71 & 37 & 33 & 36 & 34 & 28 & 64 & 39 & 40 & 57 & \textbf{47}\\
    &Joy           & 47 & 77 & 59 & 74 & 64 & 68 & 70 & 73 & \textbf{71} & 65 & 59 & 62  & 57 & 73 & 64\\
    &Neutral       & 40 & 55 & 46  & 100 & 55 & \textbf{71} & 29 & 64 & 40 & 35 & 82 & 49  & 38 & 91 & 54 \\
    &Other         & 33 & 9 & 14 & 50 & 9 & 15 & 17 & 18 & 17 & 15 & 45 & 22 & 29 & 55 & \textbf{37}  \\
    &Sadness       & 27 & 53 & 36 & 31 & 53 & 39 & 50 & 53 & \textbf{52} & 37 & 67 & 48 & 44 & 53 & 48 \\
    &Surprise      & 65 & 79 & \textbf{71} & 41 & 64 & 50 & 53 & 64 & 58  & 55 & 86 & 67  & 47 & 100 & 64   \\
    &Trust         & 39 & 69 & 50  & 86 & 46 & \textbf{60} & 67 & 31 & 42 & 43 & 77 & 56 & 50 & 62 & 55  \\
    \cmidrule(r){2-2}\cmidrule(lr){3-5}\cmidrule(rl){6-8}\cmidrule(rl){9-11}\cmidrule(rl){12-14}\cmidrule(rl){15-17}
    &Macro avg.    & 35 & 52 & 40 & 49 & 50 & 45 & 38 & 42 & 38 & 34 & 60 & 42 & 37 & 62 & \textbf{46}\\
    &Micro avg.    &    & &43 & & &45 & & &43 & & &42 & & &\textbf{47}\\
    \midrule
    \multirow{9}{*}{\rotatebox{90}{TEC}}
    &Anger        & 50 & 35 & 41     & 57 & 47 & \textbf{52} & 30 & 35 & 32
                           & 29 & 12 & 17   & 42 & 29 & 34  \\
    &Disgust      & 40 & 50 & 44   & 50 & 25 & 33 & 0 & 0 & 0   & 67 & 50 & \textbf{57}  & 50 & 50 & 50  \\
    &Fear          & 65 & 50 & 56  & 86 & 55 & \textbf{67}  & 73 & 50 & 59   & 48 & 59 & 53  & 54 & 68 & 60   \\
    &Joy            & 60 & 82 & 69  & 68 & 78 & 72 & 67 & 72 & 70   & 79 & 74 & \textbf{76}  & 66 & 82 & 73   \\
    &Sadness      & 57 & 47 & 51   & 61 & 58 & \textbf{60} & 61 & 47 & 53  & 66 & 44 & 53   & 61 & 53 & 57  \\
    &Surprise       & 48 & 32 & 39  & 45 & 50 & \textbf{47}  & 40 & 50 & 44  & 36 & 62 & 45  & 60 & 35 & 44   \\
    \cmidrule(r){2-2}\cmidrule(lr){3-5}\cmidrule(rl){6-8}\cmidrule(rl){9-11}\cmidrule(rl){12-14}\cmidrule(rl){15-17}
    &Macro avg.     & 53 & 49 & 50 & 61 & 52 & \textbf{55} & 45 & 42 & 43  & 54 & 50 & 50  & 55 & 53 & 53 \\
    &Micro avg.    & & & 57 & & & \textbf{62} & & & 56 &&& 58 &&& 61 \\
    \bottomrule
  \end{tabular}
  \captionof{table}{Performance of the neural network emotion
    classifiers. The highest F$_1$ scores are printed bold
    face.
  \label{tab:results_neural}}
\endgroup

\clearpage

\section{Neural Network Parameters}
Table \ref{tab:nn_parameters} shows the network parameters that were determined during the development process of the neural models.

\begingroup
\centering\small
\setlength{\tabcolsep}{6pt}
\renewcommand{\arraystretch}{0.9}
\begin{tabular}{ll r r r r r r}
  \toprule
  & Parameter
  & \multicolumn{1}{c}{\rotatebox{90}{\cpmNNbase}}
  & \multicolumn{1}{c}{\rotatebox{90}{\emoNNbase}}
  & \multicolumn{1}{c}{\rotatebox{90}{\emocpmNNgold}}
  & \multicolumn{1}{c}{\rotatebox{90}{\emocpmNNpred}}
  & \multicolumn{1}{c}{\rotatebox{90}{\mtlxs}}
  & \multicolumn{1}{c}{\rotatebox{90}{\mtlmh}}
  \\
  \cmidrule(r){2-2}\cmidrule(lr){3-3}\cmidrule(rl){4-4}\cmidrule(rl){5-5}\cmidrule(rl){6-6}\cmidrule(rl){7-7}\cmidrule(rl){8-8}
  \multirow{11}{*}{\rotatebox{90}{REMAN}}
  &Bi-LSTM units & 24 & 24  & 24 & 24 & 32 / 24 & 24\\
  &CNN filters & 10 & 10 & 16 & 16 & 12 / 10 & 16 \\
  &FC neurons (cpm) & 128 & ---  & 96 & 64 & 128 & 128\\
  &FC neurons (emo)  &  --- &  128 & 128 & 128 & 128 & 128 \\
  &FC neurons (comb.)  & --- & ---  & 128 & 96 & --- & --- \\
  &Loss weight (emo) & --- & 4.0 & 6.0 & 4.0 & 7.8 & 7.8\\
  &Loss weight (cpm) & 1.5 & --- & --- & --- & 1.5 & 1.5\\
  &Task weight (emo) & --- & 1.0 & 1.0 & 1.0 & 0.75 & 0.75\\
  &Task weight (cpm) & 1.0 & --- & --- & --- & 0.5 & 0.35 \\
  &Minibatch size & 60 &  50 & 50 & 50 & 25 & 25 \\
  \midrule
  \multirow{11}{*}{\rotatebox{90}{TEC}}
  &Bi-LSTM units  & 24 & 24 & 24 & 24  & 32/24 & 24\\
  &CNN filters & 32 & 32 & 32 & 32 & 24/24 & 32\\
  &FC neurons (cpm) & 32 & --- & --- & 64 & 128  & 32 \\
  &FC neurons (emo)  & ---  & 128 & 128 & 128 & 128 & 128 \\
  &FC neurons (comb.)  & --- & --- & 256 & 256& --- & --- \\
  &Loss weight (emo) & --- & 1.0 & 1.0 & 1.0& 1.0 & 1.0\\
  &Loss weight (cpm) & 1.0 & --- & --- & --- & 1.0 & 1.0 \\
  &Task weight (emo) & --- & 1.0 & 1.0 & 1.0 & 0.75 & 0.5 \\
  &Task weight (cpm) & 1.0 & --- & --- & --- & 0.5 & 0.5 \\
  &Minibatch size & 40 & 80 & 80 & 80 & 80 & 80\\
  \bottomrule
\end{tabular}
\captionof{table}{Neural network parameters. In cases where multiple values
  are displayed, the first value refers to the emotion detection
  part of the network, while the second value refers to CPM
  detection.}
\label{tab:nn_parameters}
\endgroup

\section{Discussion of Instances}

We show examples in Table \ref{tab:examples} where component
information is helpful for emotion classification.  Regarding the
neural classifiers, \mtlxs generally tends to predict fewer false
positives when there are no strong correlations among the potential
emotions to the predicted CPM, like in (1).  Similarly, in (2) the
model predicts only `fear', which is more likely to occur together
with the `subjective feeling' component than `anger' or `disgust',
according to Table 3 in the paper.  Additionally, CPM
information helps to solve ambiguities: In (3), the model predicts
`anticipation' rather than `sadness', presumably because of the
stronger correlation to the predicted CPM component `action tendency'.

In the two TEC examples (4--5), the baseline detects `joy', while \mtlxs
correctly detects `sadness'.  The cross-stitch model predicts a
`subjective feeling' component in both instances and a `cognitive
appraisal' component in one instance.  Both components are more
strongly correlated with `sadness' than with `joy' (see Table
3 in main paper).

We also show some examples that exemplify differences in prediction of
the ME-based models (6--8). Generally, the CPM information leads to
little improvement in emotion detection on TEC.  Nevertheless, there
are some cases in which the correct emotion was predicted by at least
one of \emocpmMEgold and \emocpmMEpred, whereas it was not detected by
\emoMEbase.  In both examples (6--7), the correct emotions `surprise'
and `sadness' have not been found by \emoMEbase (predicting `joy' and
`surprise' respectively).  \emocpmMEgold and \emocpmMEpred both
correctly predicted `surprise' for (6) and `sadness' for (7).  There
are indications of `subjective feeling' in the second and of `motor
expression' and `cognitive appraisal' in both examples, that were also
predicted by \cpmMEadv, which might have helped assigning the correct
emotion class. On REMAN, the ME models were able to classify a small
fraction of the instances correctly, which is still an improvement
compared to the miserably failing baseline.  An example with improved
prediction for REMAN is (8), where the emotion `joy' was correctly
identified by \emocpmMEgold and \emocpmMEpred, while not being
detected by \emoMEbase.

\vfill

\mbox{}

\begingroup
  \centering\footnotesize\scalefont{0.9}
  \begin{tabularx}{1.0\linewidth}{lX}
    \toprule
    \multicolumn{2}{p{0.97\linewidth}}{\textbf{(1)} As for the hero of this story, 'His One Fault' was
    absent-mindedness. He forgot to lock his uncle's stable
    door, and     the horse was stolen. In seeking to recover
    the stolen horse, he unintentionally stole another. (REMAN)}\\
    Emotion \emoNNbase & disgust, other, sadness \\
    CPM, \mtlxs & \textbf{cognitive appraisal} \\
    Emotion, \mtlxs & \textbf{neutral} \\
    CPM Gold & \textbf{cognitive appraisal}, action tendency \\
    Emotion Gold & \textbf{neutral} \\
    \midrule
    \multicolumn{2}{p{0.97\linewidth}}{\textbf{(2)} In that fatal valley, at the foot of that
    declivity which the cuirassiers had ascended, now inundated by the
    masses of the English, under the converging fires of the
    victorious hostile cavalry, under a frightful density of
    projectiles, this square fought on. It was commanded by an obscure
    officer named Cambronne. At each discharge, the square diminished
    and replied. (REMAN)}\\
    Emotion \emoNNbase & anger, disgust, \textbf{fear} \\
    CPM, \mtlxs & \textbf{cognitive appraisal}, subjective feeling \\
    Emotion, \mtlxs & \textbf{fear} \\
    CPM Gold & \textbf{cognitive appraisal} \\
    Emotion Gold & \textbf{fear} \\
    \midrule
    \multicolumn{2}{p{0.97\linewidth}}{\textbf{(3)} If sleep came at all, it might be a sleep
    without waking. But after all that was but one chance in a
    hundred: the action of the drug was incalculable, and the addition
    of a few drops to the regular dose would probably do no more than
    procure for her the rest she so desperately needed....  She did
    not, in truth, consider the question very closely--the physical
    craving for sleep was her only sustained sensation. Her mind
    shrank from the glare of thought as instinctively as eyes contract
    in a blaze of light--darkness, darkness was what she must have at
    any cost. (REMAN)} \\
    Emotion \emoNNbase & sadness, \textbf{fear} \\
    CPM, \mtlxs & \textbf{cognitive appraisal}, \textbf{action tendency} \\
    Emotion, \mtlxs & \textbf{fear}, \textbf{anticipation} \\
    CPM Gold & \textbf{cognitive appraisal}, neurophysiological symptoms,
               \textbf{action tendencies} \\
    Emotion Gold & \textbf{fear}, \textbf{anticipation}\\
    \midrule
    \multicolumn{2}{p{0.97\linewidth}}{\textbf{(4)} @justinbieber nocticed a girl the first
    day she got a twitter! :( (TEC)} \\
    Emotion \emoNNbase & joy \\
    CPM, \mtlxs & \textbf{cognitive appraisal}, \textbf{subjective feeling} \\
    Emotion, \mtlxs & \textbf{sadness} \\
    CPM Gold & \textbf{cognitive appraisal}, \textbf{subjective feeling} \\
    Emotion Gold & \textbf{sadness} \\
    \midrule
    \multicolumn{2}{p{0.97\linewidth}}{\textbf{(5)} when the love of your life is half way
    acrosss the world (TEC)} \\
    Emotion \emoNNbase & joy \\
    CPM, \mtlxs & \textbf{subjective feeling} \\
    Emotion, \mtlxs & \textbf{sadness} \\
    CPM Gold & \textbf{cognitive appraisal} \\
    Emotion Gold & \textbf{sadness} \\
    \midrule
    \multicolumn{2}{p{0.97\linewidth}}{\textbf{(6)} My sister is home! YAY. VISIT (TEC)}\\
    CPM \cpmMEadv & \textbf{cognitive appraisal, motor expression} \\
    Emotion \emoMEbase & joy \\
    Emotion \emocpmMEpred & \textbf{surprise} \\
    Emotion \emocpmMEgold & \textbf{surprise} \\
    CPM Gold & \textbf{cognitive appraisal, motor expression} \\
    Emotion Gold & \textbf{surprise} \\
    \midrule
    \multicolumn{2}{p{0.97\linewidth}}{\textbf{(7)} @lauren\_frost It was?!?! What the heck, man! I always miss it! Haha. - You guys need another reunion!! :) (TEC)}\\
    CPM \cpmMEadv & \textbf{cognitive appraisal, motor expression, subjective feeling} \\
    Emotion \emoMEbase & surprise \\
    Emotion \emocpmMEpred & \textbf{sadness} \\
    Emotion \emocpmMEgold & \textbf{sadness} \\
    CPM Gold & \textbf{cognitive appraisal, motor expression, subjective feeling}  \\
    Emotion Gold & \textbf{sadness} \\
    \midrule
    \multicolumn{2}{p{0.97\linewidth}}{\textbf{(8)} And if this was a necessary
    preparation for what, should follow, I would be the very last to
    complain of it. We went to bed again, and the forsaken child of some
    half-animal mother, now perhaps asleep in some filthy lodging for
    tramps, lay in my Ethelwyn's bosom. I loved her the more for it;
    though, I confess, it would have been very painful to me had she shown
    it possible for her to treat the baby otherwise, especially after what
    we had been talking about that same evening. (REMAN)}\\
    CPM \cpmMEadv & \textbf{cognitive appraisal}, action tendency, \textbf{subjective feeling}\\
    Emotion \emoMEbase & /\\
    Emotion \emocpmMEpred & \textbf{joy}  \\
    Emotion \emocpmMEgold & \textbf{joy}  \\
    CPM Gold & \textbf{cognitive appraisal, subjective feeling} \\
    Emotion Gold & disgust, \textbf{joy}, sadness, trust\\
    \bottomrule
  \end{tabularx}
  \captionof{table}{Examples in which components support emotion classification.}
  \label{tab:examples}
\endgroup

\end{document}